\documentclass[submission,copyright,creativecommons]{eptcs}
\providecommand{\event}{ICLP 2019} 
\usepackage{breakurl}             

\title{Quantified Constraint Handling Rules}
\author{Vincent Barichard
\institute{LERIA, University of Angers\\ Angers, France}
\email{vincent.barichard@univ-angers.fr}
\and
Igor St{\'e}phan
\institute{LERIA, University of Angers\\ Angers, France}
\email{igor.stephan@univ-angers.fr}
}
\def\titlerunning{Quantified Constraint Handling Rules}
\def\authorrunning{V. Barichard \& I. St{\'e}phan}

\usepackage{soul}
\usepackage{url}
\usepackage[utf8]{inputenc}
\usepackage[small]{caption}
\usepackage{graphicx}
\usepackage{color}
\usepackage{amsmath}
\usepackage{booktabs}
\usepackage{algorithm}
\usepackage{algorithmic}
\usepackage{enumitem}
\usepackage{bussproofs}
\usepackage{amsmath,amssymb,amsfonts}
\usepackage{multicol}

\def\QUACODE.{\mbox{QuaCode}}
\def\CHRPP.{\mbox{CHR++}}
\def\QCHRPP.{\ensuremath{^{Q}\mbox{CHR++}}}

\newtheorem{definition}{Definition}[section]
\newtheorem{example}{Example}[section]

\def\ILEXISTE{\ensuremath{\exists}}

\def\QUELQUESOIT{\ensuremath{\forall}}

\def\TRUE{\ensuremath{\mathit{true}}}
\def\FALSE{\ensuremath{\mathit{false}}}
\def\multiset{\mbox{multi-set}}
\def\multisets{\mbox{multi-sets}}

\def\CHRV{CHR\ensuremath{^{\vee}}}

\def\EXIST{\ensuremath{\mathit{<\!\!\exists\!\!>}}}
\def\UNIV{\ensuremath{\mathit{<\!\!\forall\!\!>}}}
\newcommand{\univ}[4]{\ensuremath{\UNIV(#1,#2,#3,#4)}}
\newcommand{\exist}[4]{\ensuremath{\EXIST(#1,#2,#3,#4)}}

\def\UpB{\ensuremath{u}}
\def\LowB{\ensuremath{l}}

\def\NIM{\ensuremath{\mathcal{N}}}
\def\BODY{\ensuremath{\Omega}}
\def\GUARD{\ensuremath{guard}}
\def\EQUALITYTHEORY{\ensuremath{ET}}

\newcommand{\SimpagationRule}[4]{{\ensuremath{#1\backslash #2 \Leftrightarrow #3\; |\; #4}}}

\newcommand{\PropagationRuleProp}[2]{{\ensuremath{#1\Rightarrow #2}}}
\newcommand{\SimpagationRuleProp}[3]{{\ensuremath{#1\backslash #2 \Leftrightarrow #3}}}
\newcommand{\SimplificationRuleProp}[2]{{\ensuremath{#1\Leftrightarrow #2}}}

\newcommand{\ExistentialSimpagationRuleProp}[6]{{\ensuremath{#1\backslash #2\EXIST[#3,#4,#5]\; #6}}}
\newcommand{\UniversalSimpagationRuleProp}[6]{{\ensuremath{#1\backslash #2\UNIV[#3,#4,#5] \;#6}}}

\newcommand{\ExistentialSimpagationRule}[7]{{\ensuremath{#1\backslash #2\EXIST[#3,#4,#5]\; #6 \;|\; #7}}}
\newcommand{\UniversalSimpagationRule}[7]{{\ensuremath{#1\backslash #2\UNIV[#3,#4,#5] \;#6\;|\; #7}}}

\def\NIMFIBO{\ensuremath{\mathit{nim\_fibo}}}
\newcommand{\nimfibo}[1]{\ensuremath{\NIMFIBO(#1)}}
\def\NIMFIBOE{\ensuremath{\mathit{nim\_fibo\_exists\_player}}}
\newcommand{\nimfiboe}[2]{\ensuremath{\NIMFIBOE(#1,#2)}}
\def\NIMFIBOU{\ensuremath{\mathit{nim\_fibo\_forall\_player}}}
\newcommand{\nimfibou}[2]{\ensuremath{\NIMFIBOU(#1,#2)}}

\def\STORE{\ensuremath{S}}
\newcommand{\seq}[4]{\ensuremath{#1~{\bf \blacktriangleright}~#2~{\bf \blacktriangleleft}~#3\vdash~#4}}

\newcommand{\eqcl}[1]{[#1]}
\def\chrprog{\ensuremath{\Gamma}}
\def\upstore{\ensuremath{\STORE_{\uparrow}}}
\def\downstore{\ensuremath{\STORE_{\downarrow}}}

\def\ETMUL{\ensuremath{\otimes}}

\def\QOMEGAL{\ensuremath{\omega^{\exists\forall}}}

\newcommand{\token}[2]{\ensuremath{#1}}

\def\GOAL{\ensuremath{\Omega}}

\def\QUELQUESOIT{\ensuremath{\forall}}
\def\ILEXISTE{\ensuremath{\exists}}
\def\BOTTOM{\ensuremath{\bot}}

\def\ET{\ensuremath{\wedge}}
\newcommand{\et}[2]{(#1 \ET #2)}

\def\EQU{\ensuremath{\leftrightarrow}}
\newcommand{\equ}[2]{(#1 \EQU #2)}
\def\IMP{\ensuremath{\rightarrow}}
\newcommand{\imp}[2]{(#1 \IMP #2)}

\def\TokenGOAL{\ensuremath{\Omega}}
\def\SubK{\ensuremath{\STORE'}}
\def\upstoreK{\ensuremath{\upstore'}}
\def\SubTokenGOAL{\ensuremath{\Omega'}}
\def\GOALB{\ensuremath{\Omega^B}}

\newcommand{\Subtree}[1]{\AxiomC{#1}}

\def\OMEGALTrue{\TRUE}

\def\TrueAxiom{\TRUE}
\newcommand{\TrueAxiomTag}{\RightLabel{\scriptsize{\TrueAxiom}}}
\newcommand{\TrueAxiomInf}[1]{\Subtree{}\TrueAxiomTag \UnaryInfC{#1}}

\def\OMEGALEquality{\emph{Equality}}
\def\EqualityRule{\ensuremath{=}}
\newcommand{\EqualityRuleTag}{\RightLabel{\scriptsize{\EqualityRule}}}
\newcommand{\EqualityRuleInf}[2]{#1\EqualityRuleTag \UnaryInfC{#2}}

\def\QOMEGALUAxiom{$\forall$-\TRUE}
\newcommand{\QOMEGALUAxiomTag}{\RightLabel{\scriptsize{\QOMEGALUAxiom}}}
\newcommand{\QOMEGALUAxiomInf}[1]{\Subtree{}\QOMEGALUAxiomTag \UnaryInfC{#1}}

\def\OMEGALTensorL{\emph{Left-elimination-of-conjunction}}
\def\TensorLRule{\ensuremath{\ETMUL_L}}
\newcommand{\TensorLRuleTag}{\RightLabel{\scriptsize{\TensorLRule}}}
\newcommand{\TensorLRuleInf}[3]{#1#2\TensorLRuleTag \BinaryInfC{#3}}

\def\QOMEGALE{$\exists$-\emph{elimination}}
\def\QOMEGALERule{\ensuremath{\exists}}
\newcommand{\QOMEGALERuleTag}{\RightLabel{\scriptsize{\QOMEGALERule}}}
\newcommand{\QOMEGALERuleInf}[2]{#1\QOMEGALERuleTag \UnaryInfC{#2}}

\def\QOMEGALU{$\forall$-\emph{elimination}}
\def\QOMEGALURule{\ensuremath{\forall}}
\newcommand{\QOMEGALURuleTag}{\RightLabel{\scriptsize{\QOMEGALURule}}}
\newcommand{\QOMEGALURuleInf}[3]{#1#2\QOMEGALURuleTag \BinaryInfC{#3}}

\def\OMEGALInactivate{\emph{Inactivate}}
\def\InactivateAxiom{\ensuremath{\uparrow}}
\newcommand{\InactivateAxiomTag}{\RightLabel{\scriptsize{\InactivateAxiom}}}
\newcommand{\InactivateAxiomInf}[1]{\AxiomC{}\InactivateAxiomTag \UnaryInfC{#1}}

\def\OMEGALApply{\emph{Apply}}
\def\OMEGALApplyRule{\ensuremath{\Leftrightarrow}}
\newcommand{\OMEGALApplyRuleTag}{\RightLabel{\scriptsize{\OMEGALApplyRule}}}
\newcommand{\OMEGALApplyRuleInf}[3]{#1#2\OMEGALApplyRuleTag\BinaryInfC{#3}}

\def\QOMEGALEApply{$\exists$-\emph{Apply}}
\def\QOMEGALEApplyRule{\ensuremath{\exists\Leftrightarrow}}
\newcommand{\QOMEGALEApplyRuleTag}{\RightLabel{\scriptsize{\QOMEGALEApplyRule}}}
\newcommand{\QOMEGALEApplyRuleInf}[2]{#1\QOMEGALEApplyRuleTag\UnaryInfC{#2}}

\def\QOMEGALUApply{$\forall$-\emph{Apply}}
\def\QOMEGALUApplyRule{\ensuremath{\forall\Leftrightarrow}}
\newcommand{\QOMEGALUApplyRuleTag}{\RightLabel{\scriptsize{\QOMEGALUApplyRule}}}
\newcommand{\QOMEGALUApplyRuleInf}[2]{#1\QOMEGALUApplyRuleTag\UnaryInfC{#2}}

\def\OMEGALSimplificationRule{\ensuremath{\Leftrightarrow}}
\newcommand{\OMEGALSimplificationRuleTag}{\RightLabel{\scriptsize{\OMEGALSimplificationRule}}}
\newcommand{\OMEGALSimplificationRuleInf}[2]{#1\OMEGALSimplificationRuleTag\UnaryInfC{#2}}

\def\tabulation{tabling}

\begin{document}
\maketitle

\begin{abstract}
We shift the QCSP  (Quantified  Constraint Satisfaction
Problems)  framework  to  the  QCHR  (Quantified  Constraint Handling Rules) framework by enabling dynamic binder and access to user-defined constraints. 
QCSP offers a natural framework to express PSPACE problems as finite two-players games. 
But to define a QCSP model, the binder must be formerly known and cannot be built dynamically even if the worst case won't occur. 
To overcome this issue, we define the new QCHR formalism that allows to build the binder dynamically during the solving.
Our QCHR models exhibit state-of-the-art performances on static binder and outperforms previous QCSP approaches when the binder is dynamic.
\end{abstract}

\section{Introduction}
We shift the QCSP (for
{\it Quantified Constraint Satisfaction Problems}) framework to the QCHR (for {\it Quantified Constraint Handling Rule}) framework. 
Such shift is motivated by the difficulties for developing real applications in QCSP.

QCSP \cite{Bordeaux_Monfroy_CP_02,Verger_Bessiere_CP_06,Gent_Nightingale_Rowley_Stergiou_AI_08,Benedetti_Lallouet_Vautard_IJCAI_07,Pralet_Verfaillie_CP_11,Mamoulis_Stergiou_CP_04,Barichard_Stephan_ICTAI_14} are a generalization of {\it Constraint Satisfaction Problems} (CSP) in which variables may be quantified existentially (as in CSP) and universally.
A QCSP is an alternation of existentially and universally quantified variables over finite domains, the binder, followed by a CSP. 
Universally quantified variables represent uncontrollable parameters such as meteorological events.
A QCSP may be seen as a two-players game in which the existentially quantified variables stand for a player $A$ and universally quantified variables stand for a player $B$.
QCSP+ \cite{Benedetti_Lallouet_Vautard_IJCAI_07} was proposed to make QCSP 
more practical from the modeling point of view.
QCSP+ uses
restricted quantification sequences instead of standard quantification sequences.
A QCSP/QCSP+ is valid if player $A$ has a strategy to win i.e. a strategy for setting the existentially quantified  variables such that no matter what setting the player $B$ chooses the CSP is true.
QCSP/QCSP+  is a rich modeling framework which leads to succinct modeling. 
But this extension also increases the complexity of solving from NP-complete to PSPACE-complete.
To fit a real problem in a QCSP/QCSP+, one has to model every part of the problem and everything must be {\it a priori} stated. 
There are problems (tic-tac-toe, reversi, connect-four) that fit this requirement, but others (checkers, chess) just do not. 
QCSP/QCSP+ cannot be used to model games and problems whose number of moves is not formerly known. 
In addition, even if everything can be statically stated, the QCSP model involves all possibles moves and overestimates the number of moves to the worst case.
For example, it is difficult to encode games such that some rules constrain future moves depending on past moves \cite{Bessiere_Verger_WMR_06}:
the QCSP will look for solutions for any possible move of player $B$
while some of them have been made impossible by previous moves.
When a QCSP/QCSP+ is defined, the solving relies on a QCSP solver \cite{Gent_Nightingale_Rowley_Stergiou_AI_08,Barichard_Stephan_ICTAI_14,Verger_Bessiere_CP_06}. 
A QCSP/QCSP+ solver is a black box program that solves a given model. 
As most QCSP/QCSP+ solvers are based on CSP solvers, there is no easy way to help the solving process by taking into account the specific properties of a quantified problem. 
In this work, we propose a new framework to model quantified problems in a dynamic way.  

CHR (for {\it Constraint Handling Rules}) \cite{Fruhwirth_TechReport_92,Fruhwirth_CP_94,Fruhwirth_JLP_98,Fruhwirth_CHR_09,Fruhwirth_Abdennadher_CHR_03,Fruhwirth_Raiser_2011} are a committed-choice language consisting of multiple-heads guarded rules that replace constraints by more  simple constraints until they are solved.
CHR are a special-purpose language concerned with defining declarative constraints in the sense of {\it Constraint logic programming} \cite{VanHentenryck_ker_91,Jaffar_Lassez_POPL_87,Jaffar_Malher_JLP_94}.
CHR are a language extension that allows to introduce {\it user-defined} constraints, i.e. first-order predicates, 
into a given host language as Prolog, Lisp, Java, or C/C++.
CHR define {\it simplification} of user-defined constraints, which replaces constraints by more simple constraints while preserving logical equivalence. 
CHR define also {\it propagation} over user-defined constraints that adds new constraints; this constraints are logically redundant but may cause further simplifications. 
CHR allow to use guards that are sequences of host language statements.
CHR finally define {\it simpagation} over user-defined constraints that mixes and subsumes simplification and propagation.
CHR (simpagation) rules are applied on \multisets\ of constraints. 
Repeated application of those rules on a \multiset\ of initial constraints incrementally solves these constraints.
The committed-choice principle expresses a \emph{don't care} nondeterminism, which leads to efficient implementations.
CHR have been extended to \CHRV\ \cite{Abdennadher_Schutz_FQAS_98} that introduces the \emph{don't know} nondeterminism in CHR \cite{Betz_Fruhwirth_TCL_13}.
This nondeterminism is freely offered when the host language is Prolog.
This nondeterminism allows to specify easily problems from the NP complexity class but it is not the case for the rest of the Polynomial Hierarchy and, in general, any problem expressed with alternating quantifications (although the formalism is Turing-complete).

We propose in this paper to extend CHR with quantification.
We call this new formalism {\it QCHR} (for \emph{Quantified Constraint Handling Rules}).
We propose to extend the simpagation rule to an \emph{existential simpagation} rule but also to a \emph{universal simpagation} rule.
The existential (resp. universal) simpagation has the same conditions to be applied as a simpagation rule but the body is existentially (resp. universally) quantified on a variable over a (finite) domain.
We obtain a formalism for which it is not necessary to declare {\it a priori} the alternation of quantifiers but where the quantifiers are generated when they are needed.
This property offers an ease of programming compared to QCSP/QCSP+ where you always need to declare a {\it finite} binder.
In QCHR, one can specify some potentially infinite games by recursion.

Section \ref{sec:motivations} presents intuitively the syntax of our QCHR formalism, illustrates on two  emblematic examples why it is not always appropriate to model and solve with QCSP/QCSP+, and shows the ease of modeling and efficiency of solving of our new formalism.
Section \ref{sec:qchr} presents our proposal, the QCHR language, with its proof-theoretical semantics.
Section \ref{sec:discussion} presents a discussion about the link between QCHR and some other related works.
Section \ref{sec:impl_exp} presents our implementation of the QCHR language into the C++ host language and some experiments.
Section \ref{sec:conclusion} concludes and draws some perspectives.
\section{Motivating examples}
\label{sec:motivations}
Our main purpose is to be able to model quantified problems when the binder can be built dynamically during the solving. 
Two well known problems are used: the Nim game and the Connect-four as motivating examples.
Informally, CHR formalism is extended with two new rules:
the existential (simpagation QCHR) rule 

\[name @\ExistentialSimpagationRule{K_1,\dots,K_m}{D_1,\dots,D_n}{it}{\LowB}{\UpB}{\GUARD}{\BODY}\]

\noindent and the universal (simpagation QCHR) rule  

\[name @\UniversalSimpagationRule{K_1,\dots,K_m}{D_1,\dots,D_n}{it}{\LowB}{\UpB}{\GUARD}{\BODY}\]

The head $(K_1,\dots,K_m \backslash D_1,\dots,D_n)$, the body \BODY\ and the \GUARD\ of these rules are interpreted in the same way as in the CHR formalism: 
Constraints $K_1,\dots,K_n$ are kept like in propagation and constraints $D_1,\dots,D_m$ are deleted like in simplification;
the constraints of the body $\BODY=B_1,\dots,B_p$ are the added constraints; if $\BODY=\TRUE$, nothing is added; if $\BODY=\FALSE$, the computation fails.
The two symbols \LowB\ and \UpB\ denotes, respectively, the lower bound and the upper bound of an integer interval.
The variable $it$ is supposed to appear in the body \BODY.
The informal semantics of those rules is as follows: 
The body \BODY\ of the existential rule leads to a success (resp. failure) if at least one value (resp. all values) $v$ taken in the interval $[\LowB..\UpB]$ leads the body $[it \leftarrow v](\BODY)$ (where the occurrences of $it$ in \BODY\ are replaced by $v$) to a success (resp. failure).
In the same way, the body \BODY\ of the universal rule leads to a success (resp. failure) if all values (resp. at least one value) $v$ taken in the interval $[\LowB..\UpB]$ leads the body $[it \leftarrow v](\BODY)$ to a success (resp. failure).
If an existential (resp. universal) QCHR rule keeps all the constraints of its multiple-head, it is an existential (resp. universal) propagation rule and is denoted by (\ExistentialSimpagationRule{K_1,\dots,K_m}{\_}{it}{\LowB}{\UpB}{\GUARD}{\BODY}) (resp. (\UniversalSimpagationRule{K_1,\dots,K_m}{\_}{it}{\LowB}{\UpB}{\GUARD}{\BODY}); 
if an existential (resp. universal) QCHR rule keeps none of the constraints of its multiple-head, it is an existential (resp. universal) simplification rule and is denoted by (\ExistentialSimpagationRule{\_}{D_1,\dots,D_n}{it}{\LowB}{\UpB}{\GUARD}{\BODY}) (resp. (\UniversalSimpagationRule{\_}{D_1,\dots,D_n}{it}{\LowB}{\UpB}{\GUARD}{\BODY}).

\paragraph{The Nim game.}
\label{sec:motivation_nimfibo}

The Nim game is a two-players game played with a heap of coins or matches.
The object of the game is to take the last match.
Each player can take one to three matches.
With the Fibonacci variant, the minimum number is one match and the maximum on the first play is one less than the initial number of matches.
Then each player may take from one to twice as many as matches as the adversary at the preceding turn.
Player $A$ begins.
For example with 4 matches, player $A$ takes 1 match. 
Then player $B$ can take 3 matches but if he does he loses immediately since he cheated.
Then player $B$ can take 1 or 2 matches.
But whatever he takes, he loses since player $A$ will take the remaining matches.
The longest possible party has 4 turns : each player takes one match at each turn.
Then QCSP has 4 quantifiers since the binder is defined statically.
With an even number $p$ of matches, the QCSP specification is as follows ($x_i$,  resp. $y_i$, is the number of matches chosen by player $A$, resp. $B$, at turn $i$): 
$R_{A}(1)$ is true (player $A$ chooses between 1 and $p-1$
matches) and for all $i$, $1 < i \leq \frac{p}{2}$, $R_{A}(i) =
(1 \leq x_{i} \leq 2 * y_{i-1}) \ET (x_i + \Sigma_{1\leq j < i} (x_j+y_j)\leq p)$ and
for all $i$, $1 \leq i \leq \frac{p}{2}$, $R_{B}(i) = (1 \leq y_{i} \leq 2 * x_{i}) \ET (\Sigma_{1\leq j \leq i} (x_j+y_j)\leq p)$ (each player takes from one to twice as many matches as the adversary at the preceding turn) \cite{Barichard_Stephan_ICTAI_14}:

\[\begin{array}{ll}
\lefteqn{\ILEXISTE x_1 \QUELQUESOIT y_1 \dots \ILEXISTE x_{\frac{p}{2}} \QUELQUESOIT y_{\frac{p}{2}} \ILEXISTE o_1\dots \ILEXISTE o_{\frac{p}{2}} }\\
&R_{A}(1) \ET \imp{R_{B}(1)}{o_1} \ET \equ{o_1}{\et{R_{A}(2)}{o_2}} \ET \\
& \equ{o_2}{\imp{R_{B}(2)}{o_3}}\ET \dots \ET \equ{o_{\frac{p}{2}}}{\imp{R_{B}(\frac{p}{2})}{\BOTTOM}}
\end{array}\]

\noindent with $x_1,y_1,\dots,x_{\frac{p}{2}},y_{\frac{p}{2}} \in [1..p-1]$ and $o_1,\ldots,o_{\frac{p}{2}}$ Boolean variables.
For each initial number of matches, one has to instantiate the general scheme.
For example, with $p=4$, the following QCSP is obtained:

\[\begin{array}{l}
\ILEXISTE x_1 \QUELQUESOIT y_1 \ILEXISTE x_2 \QUELQUESOIT y_2 \ILEXISTE o_1 \ILEXISTE o_2 \\
\imp{\et{(y_1 \leq 2*x_1)}{(x_1 + y_1 \leq 4)}}{o_1} \ET \\
\equ{o_1}{\et{\et{(x_2 \leq 2*y_1)}{(x_1 + y_1 + x_2 \leq 4)}}{o_2}} \ET \\
\equ{o_2}{\imp{\et{(y_2 \leq 2*x_2)}{(x_1 + y_1 + x_2 + y_2 \leq 4)}}{\BOTTOM}}\\
\end{array}\]

\noindent with $x_1,y_1,x_2,y_2 \in \{1,2,3\}$  and $o_1,o_2$ Boolean variables.
This QCSP is valid if the first player has a strategy for setting the existentially quantified  variables such that no matter what setting the adversary chooses for its universally quantified  variables, this first player wins (i.e. the CSP is true).

The following QCHR program solves the Nim game with the Fibonacci variant of an even or odd number of matches
($N$ represents the number of matches chosen by a player and $R$ represents the remaining number of matches into the heap):

\[\begin{array}{l}
u @  \UniversalSimpagationRuleProp{\_}{\nimfibou{N}{R}}{it}{1}{min(N,R)}{\nimfiboe{2*it}{R-it}}\\
e  @ \ExistentialSimpagationRuleProp{\_}{\nimfiboe{N}{R}}{it}{1}{min(N,R)}{\nimfibou{2*it}{R-it}}\\
\end{array}\]

These rules are built in the same way: they are existential/universal simplification rules with an omitted empty guard, the definition of the lower and upper bounds, and the body that expresses that if a player has chosen to take $it$ matches, his adversary may only choose between $1$ and $min(2*it,R)$ matches. 
The first player may freely choose between $1$ and the initial number of matches minus one:

$\begin{array}{l}
l@  \PropagationRuleProp{\nimfibo{R}}{\nimfiboe{R-1}{R}}
\end{array}$

The binder is not defined statically as for QCSP but dynamically.
Note that the base case of the recursion is hidden into the semantics of the universal rule when the lower bound becomes larger than the upper bound.

\paragraph{The Connect-four game.}
\label{sec:motivation_connectfour}
\def\ISFULLCOLUMN{\ensuremath{\mathrm{isFull}}}
\newcommand{\isfullcolumn}[1]{\ensuremath{\ISFULLCOLUMN(#1)}}
\def\SETCOIN{\ensuremath{\mathrm{coin}}}
\newcommand{\setcoin}[1]{\ensuremath{\SETCOIN(#1)}}
\def\ISGAMEWON{\ensuremath{\mathrm{isWon}}}
\newcommand{\isgamewon}[1]{\ensuremath{\ISGAMEWON(#1)}}

\def\IWU{\ensuremath{\mathit{ifRule}}}
\newcommand{\iwu}[2]{\ensuremath{\IWU(#1,#2)}}
\def\CONNECTFOURE{\ensuremath{\mathit{cfe}}}
\newcommand{\connectfoure}[1]{\ensuremath{\CONNECTFOURE(#1)}}
\def\CONNECTFOURU{\ensuremath{\mathit{cfu}}}
\newcommand{\connectfouru}[1]{\ensuremath{\CONNECTFOURU(#1)}}

The connect-four game is a two-players game in which the players first choose a color. It is played on a vertically suspended grid. At each turn, a player drops one colored coin from the top into a column of the grid. The coins fall straight down, occupying the lowest available slot within the column. The winner is the first player to form a horizontal, vertical, or diagonal line of four of one's own coins.

The connect-four game can also be modeled with a QCSP. 
In order to model every game, a grid is built for each player turn. 
There are as many game turns as there are slots in the grid. 
As a result, the QCSP model involves $\mathit{number\ of\ rows} \times \mathit{number\ of\ columns}$ grids linked with each others with constraints (see \cite{Nightingale_phd} more details about the QCSP model). 
Furthermore, constraints has to be added to detect full columns, invalid moves and winning grids.
But, even if a QCSP can be used, the binder is dynamic (a player may win the game without completely filling the board) and the model is big and not very understandable. 
In comparison, the QCHR model is more suitable, lightweight and more readable.
Let $NC$ be the number of columns of the grid. The following QCHR model computes a winning strategy if such a strategy exists:

\[\arraycolsep=1.4pt\begin{array}{rcl}
if_{\top}  & @ & \SimplificationRuleProp{\iwu{\top}{\_}}{\TRUE}\\
if_{\bot} & @ & \SimplificationRuleProp{\iwu{\bot}{N}}{\setcoin{N}, \connectfoure{\isgamewon{N}}}\\
u_{\top}  & @ & \SimplificationRuleProp{\connectfouru{\top}}{\TRUE}\\
u_{\bot} & @ & \UniversalSimpagationRuleProp{\_}{\connectfouru{\bot}}{it}{1}{NC}{\iwu{\isfullcolumn{it}}{it}}\\
e_{\top}  & @ & \SimplificationRuleProp{\connectfoure{\top}}{\FALSE}\\
e_{\bot} & @ & \ExistentialSimpagationRuleProp{\_}{\connectfoure{\bot}}{it}{1}{NC}{\setcoin{it}, \connectfouru{\isgamewon{it}}}\\
\end{array}\]

Where \setcoin{}, \isfullcolumn{} and \isgamewon{} are built-in constraints: \setcoin{} sets a coin to the given column and raises a failure if the column is full; 
\isfullcolumn{} returns $\top$ if the column is full and $\bot$ otherwise;
\isgamewon{} returns $\top$ if an alignment of four coins is found and $\bot$ otherwise. 
These functions rely on a grid $board$ which is filled according to the players moves.

The binder is not defined statically as for QCSP but dynamically. As a result, the size of the binder is equal to the number of moves done during the game and not the worst possible case.

\section{The QCHR language}
\label{sec:qchr}
A constraint is considered to be a first-order predicate.
Only one kind of predefined (built-in) constraint is required: the syntactic equality constraint denoted by $\doteq$ with an equality theory denoted  \EQUALITYTHEORY\ for variables and constants\footnote{We do not use functional symbols but the language is richer than simply variables and constants since one can use the statements of the host language that are \emph{evaluated} before equality is applied on a completely instantiated expression. It is the case for arithmetic operation in the modelling of Nim game or for more {\it ad hoc} functions like $isFull$ and $isWon$ in the modelling of the Connect-Four game.}.
For two sequences of user-defined constraints $K_1,\dots,K_m\doteq K'_1,\dots,K'_m$ means $K_1 \doteq K'_1, \dots, K_m\doteq K'_m$ and for two constraints $c(t_1,\dots,t_n)\doteq c(s_1,\dots,s_n)$ means $t_1 \doteq s_1$, \dots, $t_n\doteq s_n$.
If \STORE\ is a set of constraints, \eqcl{\STORE} denotes the set of equality constraints of \STORE.

In order to define the (proof-theoretical) semantics of QCHR, two reserved constraints are first defined.

\begin{definition}[\EXIST\ and \UNIV\ constraints]
The \EXIST\ and \UNIV\ constraints 
are constituted (in this order) of an integer variable, two integers (the lower and upper bounds) and a sequence of user-defined constraints.
\end{definition}



The \QOMEGAL\ system is based on the following kind of sequents:


\begin{definition}[\QOMEGAL\  sequent]
  \label{def:qwl_sequent}

An \QOMEGAL\ sequent is a quadruple $(\seq{\chrprog}{\TokenGOAL}{\upstore}{\downstore})$ where \downstore, the {\em down store}, and \upstore, the {\em up store}, are two stores of constraints, \chrprog\ is a sequence of CHR rules and \TokenGOAL, the {\em goal}, is a sequence  of constraints.
\end{definition}

The intuitive meaning of a sequent $(\seq{\chrprog}{\TokenGOAL}{\upstore}{\downstore})$ is  to try and consume the  constraints \TokenGOAL\ with the sequence of CHR rules \chrprog\ thanks to the store \upstore.
The elements of the store \downstore\ are the unconsumed constraints: the  constraints of \upstore\ that have not been consumed and those produced by the rules applied over \TokenGOAL\ but not consumed during this production.

The \QOMEGAL\ system is based on ten \QOMEGAL\ inference rules.
The following first five rules below are adapted from the sequent calculus system of \cite{Stephan_ICLP_18} for CHR.
This implies that a QCHR program may contain CHR rules.
In the following, some constraints ($A$, $K_1,\dots,K_m$, $D_1,\dots,D_n$),
some stores of constraints (\STORE, $\STORE^K$, $\STORE^D$, \SubK, \upstoreK, \ldots) and some sequences of constraints (\TokenGOAL, \SubTokenGOAL) are used.

\paragraph{The \OMEGALApply\ inference rule:}

\begin{prooftree}
\OMEGALApplyRuleInf
{\AxiomC{\seq{\chrprog}{\Omega^B}{\STORE^K,\STORE^O,\STORE^{K}\doteq\STORE^{K'},\STORE^{D}\doteq\STORE^{D'}}{\SubK,\STORE''}}}
{\AxiomC{\seq{\chrprog}{\SubTokenGOAL}{\STORE''}{\downstore}}}
{\seq{\chrprog}{A}{\STORE^D,\STORE^K,\STORE^O}{\downstore}}
\end{prooftree}

with 
\begin{itemize}
\item  (\SimpagationRule{K'_1, \dots, K'_m}{D'_1, \dots, D'_n}{\GUARD}{\GOALB}) a rule of \chrprog;
\item $\STORE^{K} = \{{K_1},\dots, {K_m}\}$, $\STORE^{D}= \{{D_1},\dots, {D_n}\}$ and there exists $j$ such that
\begin{itemize}
\item either $1\leq j \leq n$, $D_j=A$, 
\item or $1\leq j \leq m$, $K_j=A$; 
\end{itemize}
\item $\EQUALITYTHEORY\models\imp{\eqcl{\STORE^O}}{\exists\overline{X}(\GUARD \ET (\STORE^{K}\doteq\STORE^{K'}) \ET (\STORE^{D}\doteq\STORE^{D'}))}$  ($\overline{X}$ the set of variables of $\STORE^{K'} = \{{K'_1},\dots, {K'_m}\}$ and $\STORE^{D'}= \{ {D'_1},\dots, {D'_n}\}$);
\item \SubTokenGOAL\ is a sequence composed of all the elements of $\SubK\subseteq \STORE^{K}$.
\end{itemize}

The \OMEGALApply\ inference rule applies a QCHR rule on a constraint $A$ since there are two sub-stores $\STORE^K$ and $\STORE^D$ of the store $\upstore = \STORE^K\uplus\STORE^D\uplus\STORE^O$ such that $\STORE^K\uplus\STORE^D\uplus\{A\} =  \{{K_1},\dots, {K_m},$ ${D_1},\dots, {D_n}\}$ modulo the equality constraints of 
$\STORE^O$ and such that the sequence of host statements and equalities of the guard, \GUARD, are verified.
The solving of the constraint $A$ is reduced to the solving of the goal $\Omega^B=B_1,\dots,B_p$ of the CHR rule  and eventually the solving of the constraints of $\Omega'$ in the case that constraints from  $\SubK\subseteq \STORE^K$  were not consumed during the process of consumption/production of \GOALB.
A part of the resources 
$\STORE^O$ is allocated to solve the goal \GOALB, the rest of the constraints  and those produced by \GOALB\ but unconsumed, $\STORE''$, are allocated to a sequence \SubTokenGOAL\ over \SubK.
Since the \QOMEGAL\ system only applies a QCHR rule if one of the constraints of its head is focused on, the calculus of (\seq{\chrprog}{\SubTokenGOAL}{\STORE''}{\downstore}) is necessary to the completeness of \QOMEGAL\ w.r.t. the semantics of CHR for non-quantified CHR programs (see Example \ref{ex:apply_rule_1} where $\SubK=\{a\}$).
But, \SubK\ may be empty if all the resources have been consumed
(see Example \ref{ex:apply_rule_3} where $\SubK$ is empty).
If the applied rule is a simplification rule (i.e. $\STORE^{K}$ is empty) or \SubK\ is empty then \SubTokenGOAL\ is empty and the right above sequent is omitted (and $\downstore = \STORE''$).
The \OMEGALApply\ inference rule realizes in fact a hidden use of the cut-rule of the linear-logic sequent calculus \cite{Girard_TCS_87}: A lemma is computed by the left sub-proof and used in the right sub-proof\footnote{see \cite{Stephan_ICLP_18} for a discussion about the linear-logic properties of this rule}.

\paragraph{The \OMEGALTensorL\ inference rule:}

\begin{prooftree}
\TensorLRuleInf
{\Subtree{\seq{\chrprog}{A}{\upstore}{\STORE^{O}}}}
{\Subtree{\seq{\chrprog}{\TokenGOAL}{\STORE^{O}}{\downstore}}}
{\seq{\chrprog}{A, \TokenGOAL}{\upstore}{\downstore}}
\end{prooftree}

If the current goal is a sequence of constraints, the \OMEGALTensorL\ inference rule is applied: 
The first constraint $A$ of the sequence is isolated and a part of the resources \upstore\ are allocated to solve the constraint; the rest of the constraints and those produced by $A$ but unconsumed, $\STORE^{O}$, are allocated to the remaining sequence of constraints.
This inference rule realizes also, in fact, a hidden use of the cut-rule of the linear-logic sequent calculus: The \downstore\ is a lemma computed by the left sub-proof and used in the right sub-proof.


\paragraph{The \OMEGALInactivate\ axiom:}

\begin{prooftree}
\InactivateAxiomInf
{\seq{\chrprog}{A}{\STORE}{A,\STORE}}
\end{prooftree}

\begin{sloppypar}
with no QCHR rule (\SimpagationRule{K'_1, \dots, K'_m}{D'_1, \dots, D'_n}{\GUARD}{\BODY}) of \chrprog\ such that $j$, 
($1\leq j \leq n$, $D_j=A$ or 
$1\leq j \leq m$, $K_j=A$), 
 $\STORE^D=\{D_1,\dots,D_n\}\subseteq \STORE$, $\STORE^K=\{K_1,\dots,K_m\}\subseteq \STORE \setminus \STORE^D$;
and $\EQUALITYTHEORY\models \imp{\eqcl{\STORE}}{\exists\overline{X}(\GUARD \wedge (S^K \doteq S^{K'}) \wedge (S^D \doteq S^{D'}))}$ ($\overline{X}$ the set of variables of $\STORE^{K'} =\{ {K'_1},\dots, {K'_m}\}$ and $\STORE^{D'}= \{{D'_1},\dots, {D'_n}\}$).
\end{sloppypar}

If there is no QCHR rule to consume the current user-defined constraint $A$ by an apply rule, the \OMEGALInactivate\ axiom stores the constraint into the store.



\paragraph{The \OMEGALTrue\ axiom:}
\begin{prooftree}
\TrueAxiomInf{\seq{\chrprog}{\TRUE}{\STORE}{\STORE}}
\end{prooftree}

If the current goal is the \TRUE\ constraint then no constraint is consumed and the \OMEGALTrue\ axiom is applied.

\paragraph{The \OMEGALEquality\ inference rule:}
\begin{prooftree}
\EqualityRuleInf
{\Subtree{\seq{\chrprog}{\GOAL^{X,Y}}{\STORE^O, (X \doteq Y)}{\downstore}}}
{\seq{\chrprog}{(X \doteq Y)}{\STORE^O, \STORE^{X,Y}}{\downstore}}
\end{prooftree}
with $\{\STORE^O, \STORE^{X,Y}\}$  a partition of the store such that $\STORE^{X,Y}$ is the set of identified user-defined constraints that contain either the variables $X$ or $Y$, $\GOAL^{X,Y}$ a sequence over $\STORE^{X,Y}$, and 
with the proviso that the equality constraint ($X \doteq Y$) is consistent with the equivalence classes \eqcl{$\STORE^O$} according to \EQUALITYTHEORY. 

If the current goal is only an equality constraint $(X \doteq Y)$ then no constraint is consumed and the \OMEGALEquality\ inference rule is applied:
The equality constraint is added to the store of constraints $\STORE^O$ and a sequence  $\GOAL^{X,Y}$ of constraints, over all the constraints  $\STORE^{X,Y}$ of the store with occurrences of variables $X$ or $Y$, is inserted into the goal part of a sequent.
Since equivalence classes for the variables are modified, some rules might be applied from now on.



\begin{example}
\label{ex:apply_rule}
~
\begin{enumerate}[label=\textbf{\arabic*},ref=\ref{ex:apply_rule}.\arabic*]
\item \label{ex:apply_rule_1} Let \chrprog\ be the CHR program
(\SimpagationRuleProp{a}{b}{\TRUE}), (\SimplificationRuleProp{a,c}{\TRUE}) and the goal $b,c,a$ then 

{\small
\begin{prooftree}
\OMEGALApplyRuleInf
{\TrueAxiomInf{\seq{\chrprog}{\TRUE}{a,c}{a,c}}}
{
\OMEGALSimplificationRuleInf
{\TrueAxiomInf{\seq{\chrprog}{\TRUE}{}{}}}
{\seq{\chrprog}{a}{c}{}
}
}
{\seq{\chrprog}{a}{b,c}{}
}
\end{prooftree}
}


\item \label{ex:apply_rule_3} Let \chrprog\ be the CHR program (\SimpagationRuleProp{a,b}{c}{d}), (\SimplificationRuleProp{a,b,d}{\TRUE}) and the goal $a,b,c$ then 

{\small
\begin{prooftree}
\OMEGALSimplificationRuleInf
{\OMEGALSimplificationRuleInf
{\TrueAxiomInf{\seq{\chrprog}{\TRUE}{}{}}}
{\seq{\chrprog}{d}{a,b}{}}}
{\seq{\chrprog}{c}{a,b}{}
}
\end{prooftree}
}

\end{enumerate}
\end{example}
Below, the five rules that are specific to manage the quantifiers are described.

\paragraph{The \QOMEGALEApply\ inference rule:}

\begin{prooftree}
\QOMEGALEApplyRuleInf
{\Subtree{\seq{\chrprog}{\exist{it}{\LowB}{\UpB}{\GOALB}}{\STORE^K,\STORE, \STORE^{K}\doteq\STORE^{K'},\STORE^{D}\doteq\STORE^{D'}}{\downstore}}}
{\seq{\chrprog}{A}{\STORE^D,\STORE^K,\STORE}{\STORE^K,\STORE}}
\end{prooftree}

with 
\begin{itemize}
\item (\ExistentialSimpagationRule{K'_1, \dots, K'_m}{D'_1, \dots, D'_n}{it}{\LowB}{\UpB}{\GUARD}{\GOALB}) is a QCHR rule of \chrprog;
\item $\STORE^{K} = \{{K_1},\dots, {K_m}\}$, $\STORE^{D}= \{{D_1},\dots, {D_n}\}$ and there exists $j$ such that
\begin{itemize}
\item either $1\leq j \leq n$, $D_j=A$, 
\item or $1\leq j \leq m$, $K_j=A$; 
\end{itemize}
\item $\EQUALITYTHEORY\models\imp{\eqcl{\STORE}}{\exists\overline{X}(\GUARD \ET (\STORE^{K}\doteq\STORE^{K'}) \ET (\STORE^{D}\doteq\STORE^{D'}))}$  ($\overline{X}$ the set of variables of $\STORE^{K'} = \{{K'_1},\dots, {K'_m}\}$ and $\STORE^{D'}= \{ {D'_1},\dots, {D'_n}\}$).
\end{itemize}

The \QOMEGALEApply\ inference rule applies, in the same conditions as an \OMEGALApply\ rule, an existential QCHR rule but introduces only one constraint: an existential constraint
since the elimination of a quantified constraint is part of the mechanism of the semantics.



\paragraph{The \QOMEGALUApply\ inference rule:}

\begin{prooftree}
\QOMEGALUApplyRuleInf
{\Subtree{\seq{\chrprog}{\univ{it}{\LowB}{\UpB}{\GOALB}}{\STORE^K,\STORE,\STORE^{K}\doteq\STORE^{K'},\STORE^{D}\doteq\STORE^{D'}}{\downstore}}}
{\seq{\chrprog}{A}{\STORE^D,\STORE^K,\STORE}{\STORE^K,\STORE}}
\end{prooftree}

with 
\begin{itemize}
\item (\UniversalSimpagationRule{K'_1, \dots, K'_m}{D'_1, \dots, D'_n}{it}{\LowB}{\UpB}{\GUARD}{\GOALB}) is a QCHR rule of \chrprog;

\item $\STORE^{K} = \{{K_1},\dots, {K_m}\}$, $\STORE^{D}= \{{D_1},\dots, {D_n}\}$ and there exists $j$ such that
\begin{itemize}
\item either $1\leq j \leq n$, $D_j=A$, 
\item or $1\leq j \leq m$, $K_j=A$; 
\end{itemize}
\item $\EQUALITYTHEORY\models\imp{\eqcl{\STORE}}{\exists\overline{X}(\GUARD \ET (\STORE^{K}\doteq\STORE^{K'}) \ET (\STORE^{D}\doteq\STORE^{D'}))}$  ($\overline{X}$ the set of variables of $\STORE^{K'} = \{{K'_1},\dots, {K'_m}\}$ and $\STORE^{D'}= \{ {D'_1},\dots, {D'_n}\}$).
\end{itemize}

In the same way, the \QOMEGALUApply\  inference rule applies, in the same conditions as an \OMEGALApply\ rule, an universal QCHR rule but introduces only one constraint: an universal constraint.



\paragraph{The \QOMEGALE\ inference rule:}

\begin{prooftree}
\QOMEGALERuleInf
{\Subtree{\seq{\chrprog}{\TokenGOAL^{it}}{\STORE}{\downstore}}}
{\seq{\chrprog}{\exist{it}{\LowB}{\UpB}{(B_1,\dots,B_p)}}{\STORE}{\STORE}}
\end{prooftree}\label{par:qomegale}

with  $\LowB \leq \UpB$, $x\in [\LowB .. \UpB]$ and $\TokenGOAL^{it} = \token{[it \leftarrow x](B_1)}{i'},\dots, \token{[it \leftarrow x](B_p)}{i'+p}$.

If the goal is a unique existential constraint, then the \QOMEGALE\ inference rule is applied: A value $x$ into the interval $[\LowB..\UpB]$ is chosen and assigned to the variable $it$ of the constraints $B_1$, \dots, $B_p$ (this is the meaning of $[it \leftarrow x](B_i), 1\leq i\leq p$).
If $\LowB > \UpB$ then, there is no possible sub-proof from this sequent.
The resulting store \downstore\ is ignored after the proof of the $\TokenGOAL^{it}$: production and consumption are local.



\paragraph{ The \QOMEGALU\ inference rule:}\label{par:qomegalu}
\begin{prooftree}
\QOMEGALURuleInf
{\Subtree{\seq{\chrprog}{\TokenGOAL^{it}}{\STORE}{\STORE^{\LowB}}}}
{\Subtree{\seq{\chrprog}{U^{l+1}}{\STORE}{\STORE^{\LowB+1}}}}
{\seq{\chrprog}{\univ{it}{\LowB}{\UpB}{(B_1,\dots,B_p)}}{\STORE}{\STORE}}
\end{prooftree}

with $\LowB \leq \UpB$ and 
$U^{l+1}=\univ{it}{\LowB+1}{\UpB}{(B_1,\dots,B_p)}$,  $\TokenGOAL^{it} = [it \leftarrow \LowB](B_1),\dots, [it \leftarrow \LowB](B_p)$.

If the goal is a unique universal constraint, then the \QOMEGALU\ inference rule is applied and leads to two sub-proof trees: The lower bound of the interval $[\LowB..\UpB]$, $\LowB \leq \UpB$, is assigned to the variable $it$ of the constraints $B_1$, \dots, $B_p$ and the sequent (\seq{\chrprog}{[it \leftarrow \LowB](B_1),\dots, [it \leftarrow \LowB](B_p)}{\STORE}{\STORE^{\LowB}}) has to be proved;
and the lower bound is increased by 1 and the sequent (\seq{\chrprog}{\univ{it}{\LowB+1}{\UpB}{(B_1,\dots,B_p)}}{\STORE}{\STORE^{\LowB+1}}) has also to be proved.
The resulting stores $\STORE^{\LowB}$ and $\STORE^{\LowB+1}$ are ignored after the proofs of $\TokenGOAL^{it}$ and $U^{l+1}$: production and consumption are local.

\paragraph{The \QOMEGALUAxiom\ axiom:}

\begin{prooftree}
\QOMEGALUAxiomInf{\seq{\chrprog}{\univ{it}{\LowB}{\UpB}{B}}{\STORE}{\STORE}}
\end{prooftree}

with $\UpB < \LowB$.

Finally, if the goal is a unique universal constraint with $\UpB < \LowB$, then the \QOMEGALUAxiom\ axiom is applied since the logic formula $\imp{x\in [\LowB..\UpB]}{B}$ is equivalent to \TRUE.

Now, the \QOMEGAL\ sequent calculus is defined:

\begin{definition}[\QOMEGAL\ sequent calculus system]
  \label{def:qwl_system}
  The \QOMEGAL\ sequent calculus system is the given of the ten previous inference rules and axioms.
 \end{definition}

Restricted to the first four items (ie. \OMEGALApply, \OMEGALTensorL\ and \OMEGALInactivate\ inference rules and \OMEGALTrue\ axiom), the system is equivalent to the proof-theoretical semantics of CHR of  \cite{Stephan_ICLP_18} lifted to first order thanks to the \OMEGALEquality\ inference rule.

\begin{example}

The QCHR program \NIM\ that solves the Nim game, one of the motivating examples of Section \ref{sec:motivations}, is recalled
($N$ represents the number of matches chosen by a player and $R$ represents the remaining number of matches into the heap):

\[\begin{array}{l}
u @  \UniversalSimpagationRuleProp{\_}{\nimfibou{N}{R}}{it}{1}{min(N,R)}{\nimfiboe{2*it}{R-it}}\\
e  @ \ExistentialSimpagationRuleProp{\_}{\nimfiboe{N}{R}}{it}{1}{min(N,R)}{\nimfibou{2*it}{R-it}}\\
\end{array}\]

and a proof of (\seq{\NIM}{\nimfibo{4}}{}{}) is given:

\begin{prooftree}
\OMEGALSimplificationRuleInf{
\QOMEGALEApplyRuleInf{
\QOMEGALERuleInf{
\QOMEGALUApplyRuleInf{
\QOMEGALURuleInf{
\QOMEGALEApplyRuleInf{
\QOMEGALERuleInf{
\QOMEGALUApplyRuleInf{
\QOMEGALUAxiomInf{\seq{\NIM}{\univ{i}{1}{0}{\nimfiboe{2*i}{0-i}}}{}{}}}
{\seq{\NIM}{\nimfibou{4}{0}}{}{}}}
{\seq{\NIM}{\exist{i}{1}{2}{\nimfibou{2*i}{2-i}}}{}{}}}
{\seq{\NIM}{\nimfiboe{2}{2}}{}{}}}
{\Subtree{$\nabla$}}
{\seq{\NIM}{\univ{i}{1}{2}{\nimfiboe{2*i}{3-i}}}{}{}}}
{\seq{\NIM}{\nimfibou{2}{3}}{}{}}}
{\seq{\NIM}{\exist{i}{1}{3}{\nimfibou{2*i}{4-i}}}{}{}}}
{\seq{\NIM}{\nimfiboe{3}{4}}{}{}}}
{\seq{\NIM}{\nimfibo{4}}{}{}}
\end{prooftree}

with $\nabla$:

\begin{prooftree}
\QOMEGALURuleInf
{\QOMEGALEApplyRuleInf{
\QOMEGALERuleInf{
\QOMEGALUApplyRuleInf{
\QOMEGALUAxiomInf{\seq{\NIM}{\univ{i}{1}{0}{\nimfiboe{2*i}{0-i}}}{}{}}}
{\seq{\NIM}{\nimfibou{2}{0}}{}{}}}
{\seq{\NIM}{\exist{i}{1}{1}{\nimfibou{2*i}{1-i}}}{}{}}}
{\seq{\NIM}{\nimfiboe{4}{1}}{}{}}}
{\Subtree{$\nabla'$}}
{\seq{\NIM}{\univ{i}{2}{2}{\nimfiboe{2*i}{3-i}}}{}{}}
\end{prooftree}

and $\nabla'$:

\begin{prooftree}
\QOMEGALUAxiomInf{\seq{\NIM}{\univ{i}{3}{2}{\nimfiboe{2*i}{3-i}}}{}{}}
\end{prooftree}
\end{example}

\section{Discussion}
\label{sec:discussion}
All the state-of the-art QCSP solvers have the same drawback: they explore much larger combinatorial spaces than the natural search space of the original problem.
In \cite{Barichard_Stephan_ICTAI_14}, the meaning of the ``Achilles'heel'' notion, initially introduced for Quantified Boolean Formulas (QBF) \cite{Cadoli_Giovanardi_Schaerf_AAAI_98,Rabe_Tentrup_FMCAD_15,Zhang_AAAI_06,Zhang_Malik_ICCAD_02} in \cite{Ansotegui_Gomes_Selman_AAAI_05}  as the difficulty to detect that the Boolean constraints are necessarily true under some partial assignment,
has been extended for QCSP
to the larger problem of how to avoid the exploration of combinatorial spaces that are known to be useless by construction.
This definition includes the capture of the illegal actions of the player $B$ but also for example the end of the game before the last turn that is also a source of oversized explored search space.
Already cited, \cite{Benedetti_Lallouet_Vautard_IJCAI_07,Verger_Bessiere_CP_08} proposes the new {\it QCSP+} language that use restricted quantification sequences instead of standard quantification sequences.
Some other approaches propose to modify more or less the QCSP/QCSP+ language to overcome the different drawbacks:  
\cite{Bessiere_Verger_WMR_06} proposes the new {\it Strategic CSP} language where universal variables adapt their domain to be compatible with previous choices; 
\cite{Pralet_Verfaillie_CP_11} proposes also a new language but restricted to Markovian Game-CSP, based explicitly on the notion of state, in order to efficiently model and solve control problems for completely observable and Markovian dynamic systems.
Those language are easily and efficiently expressed in QCHR (see Subsection \ref{subsec:Nim_exp} "The Nim Game" to a discussion about states in QCHR to improve efficiency).

\section{Implementation and Experiments}
\label{sec:impl_exp}
To evaluate the QCHR approach, three well known problems\footnote{We used the matrix game, the nimfibo game and the connect-four game to carry our experiments.} are modelled and solved in QCSP and QCHR. 
Then, the results obtained by two solvers are compared: \QUACODE. \cite{Barichard_Stephan_ICTAI_14} for the solving of the QCSP models and our solver \QCHRPP. for the solving of the QCHR models.
\QCHRPP. is a QCHR solver based on the \CHRPP. solver\footnote{\CHRPP. is a CHR solver built on the top of the C++ language, it can be downloaded at \url{https://gitlab.com/vynce/chrpp}}. The latter has been extended by adding two new items to its grammar: \texttt{exists} and \texttt{forall}. Item \texttt{exists} allows you to browse the domain of a variable by successively trying each of its values in search of success. Item \texttt{forall} makes it possible to browse the domain of a variable by making sure that each of the values leads to a success. They implement the \QOMEGALE\ and \QOMEGALU\ rules previously defined.

Each experiment has been run $10$ times and the average of the running times and number of failures are reported in the tables. 
Notice that there is no random parameter or value used for both solvers. 
As a consequence, for a given instance, the number of failures is always the same and the standard deviation of the running times is quite low.  
All experiments have been run on an Intel Xeon-E5, 2.1-3.3GHz, 128GB RAM running Linux. 
Maximum computation time is set to $300$ seconds. All benchmarks used here are provided with CHR++ sources.
\def\MATRIXGAMEE{\ensuremath{\mathit{mge}}}
\newcommand{\matrixgamee}[1]{\ensuremath{\MATRIXGAMEE(#1)}}
\def\MATRIXGAMEU{\ensuremath{\mathit{mgu}}}
\newcommand{\matrixgameu}[1]{\ensuremath{\MATRIXGAMEU(#1)}}

\def\UPDATECORNERU{\ensuremath{\mathrm{updateCorner_\forall}}}
\newcommand{\updatecorneru}[1]{\ensuremath{\UPDATECORNERU(#1)}}
\def\UPDATECORNERE{\ensuremath{\mathrm{updateCorner_\exists}}}
\newcommand{\updatecornere}[1]{\ensuremath{\UPDATECORNERE(#1)}}

\paragraph{The matrix game}
\label{subsec:matrix_exp}
is a two-players game of $d$ turns. 
It is played on a $0/1$ square matrix of size $2^d$. 
At each turn, player $A$ cuts the matrix in half horizontally and decides to keep the top or bottom part. 
Player $B$ then cuts the matrix in half vertically and keeps the left or right part. 
If the last cell contains a $1$, player $A$ wins.

The matrix game is well suited to the QCSP approach because the binder does not take benefit from being defined dynamically. 
Indeed, the number of moves of a matrix game is fixed and only depends on the matrix size. 
Although it does not belong to our motivating examples, it is used as a use case to evaluate the efficiency of our approach on static binders.
Let $M$ be the input matrix of size $2^d$ such that $M[i,j] \in \{0,1\}, i,j \in [0;\sqrt{2^d}-1]$. 
The following QCHR program finds a winning strategy to the matrix game:

\[\arraycolsep=1.4pt\begin{array}{rcl}
u   & @ & \UniversalSimpagationRuleProp{\_}{\matrixgameu{d}}{it}{0}{1}{\updatecorneru{it}, \matrixgamee{d-1}}\\
e_0 & @ & \SimplificationRuleProp{\matrixgamee{0}}{M[CornerUL_x,CornerUL_y] = 1}\\
e   & @ & \ExistentialSimpagationRuleProp{\_}{\matrixgamee{d}}{it}{0}{1}{\updatecornere{it}, \matrixgameu{d-1}}\\
\end{array}\]

With $CornerUL$ and $CornerLR$ the coordinates of the upper left and lower right corners of the relevant part of the matrix. 
The relevant part is the remaining part after a player turn.
Constraints \updatecorneru{} and \updatecornere{} are built-in constraints that update the upper left and lower right corners of the relevant part of the matrix. 
The initial call is \matrixgamee{2*d}.

The matrix game has already been modeled as a QCSP and solved with a QCSP solver. 
Random matrices are used as input for the instances and computed the average execution time and number of failures of each solver. 
Results are reported in Table~\ref{table:Res_MatrixGame}. 



\begin{table}
\scriptsize
\centering
\begin{tabular}{|r|r|r|r|r|}
\hline
Depth  & \multicolumn{2}{c|}{\QUACODE.} &  \multicolumn{2}{c|}{\QCHRPP.} \\
\cline{2-5}
& Failures  & Time (s) & Failures & Time (s)\\
\hline
4 & 14 & $<$1 & 22 & $<$1 \\
6 & 40 & $<$1 & 50 & $<$1 \\
8 & 235 & 3.9 & 366 & $<$1 \\
9 & 447 & 59.53 & 705 & $<$1 \\
10 & TL & TL & 1340 & $<$1 \\
\hline
\end{tabular}
\caption{Results for the Matrix game\label{table:Res_MatrixGame}}
\end{table}

Both models are quite equivalent and the solving takes no benefit from propagation of QCSP constraints.
The number of failures encountered by \QUACODE. is smaller than that of \QCHRPP..
Indeed, the propagation done in \QUACODE. prevents the last branching steps of search. 
But the cost is huge compared to the price of propagating the previous choices during the whole search. 
In comparison, \QCHRPP. does not propagate the choices thanks to the constraints but its model is quite simple to test and each basic operation is constant in time. 
Indeed, it only has to update the corner variables and to check at the end if it is a $1$ or a $0$. 
That is why \QCHRPP. is able to solve the instance of depth $10$ while \QUACODE. reaches the time limit.
As a result, we show that the QCHR model combined with \QCHRPP. is lighter and can be more efficient than a QCSP model with \QUACODE..
\paragraph{The Nim game}
\label{subsec:Nim_exp}
is our first motivating example which has been introduced in Section \ref{sec:motivation_nimfibo}. 
It can be modeled in QCHR in a very intuitive way.
But the QCHR model can do even more. 
Indeed, thanks to its dynamic way of modeling, it is possible to make a top-down evaluation with \tabulation\ \cite{Schrijvers_Warren_ICLP_04}. 
A top-down evaluation with \tabulation\ consists in storing states of sub-trees already explored. 
If such a sub-tree is encountered another time, its result will be used instead of exploring it again.


\begin{figure}
    \centering
    \resizebox{0.8\linewidth}{!}{\includegraphics{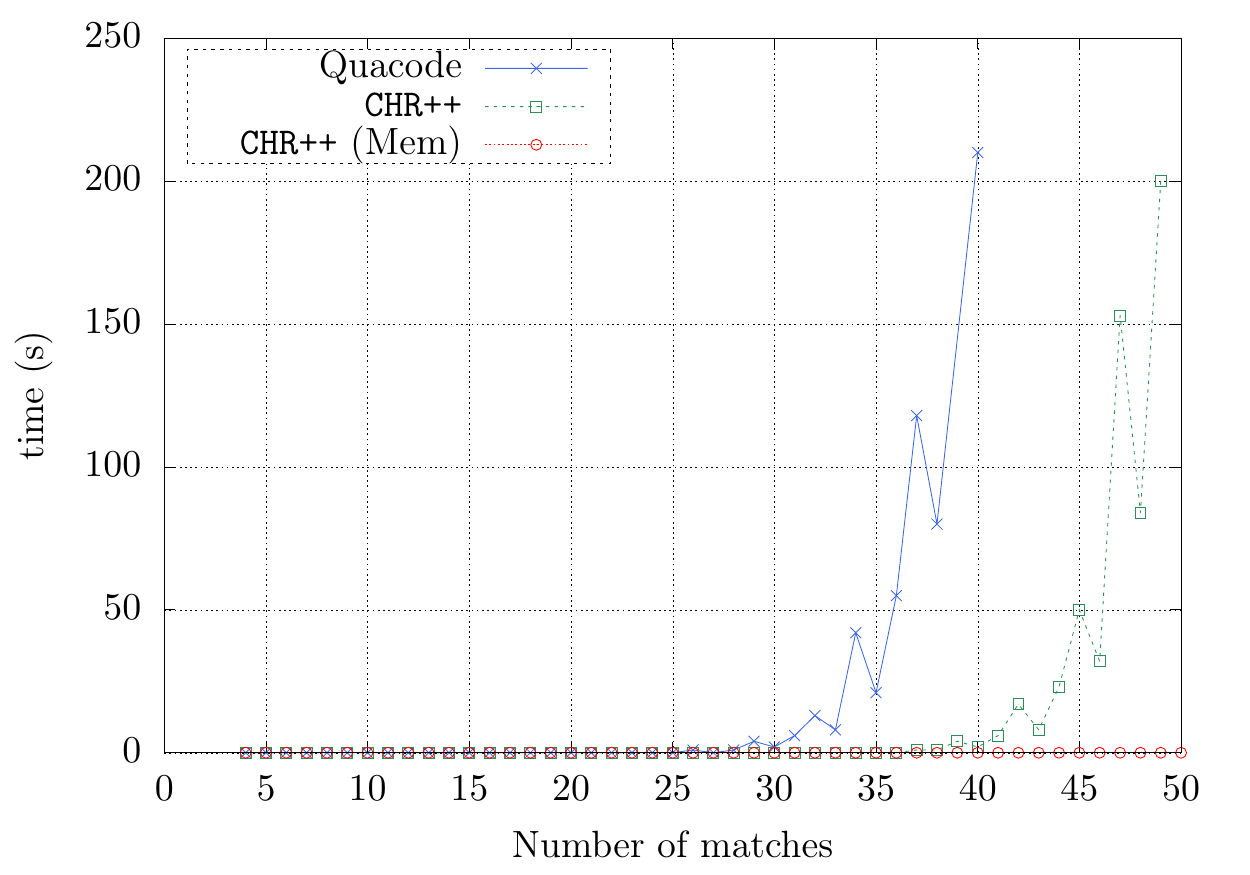}}
    \caption{Nim execution time}
    \label{fig:res_nimfibo_runtime}
\end{figure}

For the experiments, we compare three solving approaches: \emph{\QUACODE.} for the QCSP/QCSP+ solving, \emph{\QCHRPP.} for the QCHR solving and \emph{\QCHRPP. (Mem)} for the solving of the QCHR model with \tabulation.
We first observe on Figure \ref{fig:res_nimfibo_runtime} that \QUACODE. cannot solve instances of more than $40$ matches without exceeding the time limit (i.e. $300s$). 
We notice that \QUACODE. and \QCHRPP. encountered the same number of failures. 
Indeed, they are both based on same model and there is no constraint to propagate. 
But, as seen on Figure \ref{fig:res_nimfibo_runtime}, \QCHRPP. performs better. 
We now compare with \QCHRPP. with \tabulation\ and see on Figure \ref{fig:res_nimfibo_runtime} that all instances are solved in less than $1$ second. 
This huge improvement is explained by the \tabulation\ process. Indeed, a Nim game state\footnote{A Nim game state is given by the player number, the number of matches} can be encountered many times during the search. 
Recording such states will prevent the algorithm to explore them again in the future and increases a lot the efficiency of the search. 
As the current state of a Nim game is Markovian (i.e. the current state does not depend on the whole history of events), similar results are achieved in \cite{Pralet_Verfaillie_CP_11}.

\paragraph{The connect-four game}
\label{subsec:connect-four_exp}
is our second motivating example. It has been presented at Section \ref{sec:motivation_connectfour}. The QCSP+ model for the connect-four is big and not very understandable compared to the QCHR model. The QCHR model involves a dynamic binder which is built during the solving.

\begin{table}
\scriptsize
\centering
\begin{tabular}{|c|c|r|r|r|r|}
\hline
\multicolumn{2}{|c|}{Board size}  & \multicolumn{2}{c|}{\QUACODE.} &  \multicolumn{2}{c|}{\QCHRPP.} \\
\cline{3-6}
\multicolumn{2}{|c|}{}  & Failures  & Time (s)  & Failures & Time (s)\\
\hline
~4~~ & 4 & 2123 & $<$1 & 28818 & $<$1 \\
~4~~ & 5 & 26754 & 9.53 & 327561 & $<$1 \\
~5~~ & 4 & 312580 & 105.75 & 5373028 & 2.17 \\
~5~~ & 5 & TL & TL & 120470758 & 75.83 \\
\hline
\end{tabular}
\caption{Connect-four\label{table:Res_ConnectFour}}
\end{table}

For the experiments, \emph{\QUACODE.} for the QCSP solving and \emph{\QCHRPP.} for the QCHR solving are compared. 
As shown Table \ref{table:Res_ConnectFour}, the number of failures encountered by \QUACODE. is smaller than that of \QCHRPP.. 
Indeed, as for the matrix game, the propagation done in \QUACODE. prevents the last branching steps of search. 
But the cost is huge compared to the benefit of avoiding a few branching steps. 
The results show that \QCHRPP. performs better than \QUACODE.. 
It is more than $50$ times faster and can even find solution before the time limit on the last instance. For larger boards, computing times become enormous, exceeding the maximum allowed time.

\section{Conclusion}
\label{sec:conclusion}
This paper proposed the new \emph{QCHR} formalism which is an extension of CHR with quantification. 
QCHR allows to model dynamic binders. 
This overcomes one of the main drawbacks of the QCSP framework.
We also presented some very intuitive \emph{QCHR} models and we solved them with a \emph{QCHR} solver, \QCHRPP..
The experiments showed that \QCHRPP. always outperforms the QCSP solver and sometimes by many orders of magnitude.

We believe that this new formalism gives an easier way to model constrained problems with quantifications and offers a new way when the binder cannot be statically deduced. 
In the future, we plan to tackle problems that cannot be dealt with QCSP such as problems where a static binder cannot be found.

\bibliographystyle{eptcs}
\bibliography{references}
\end{document}